\documentclass[review]{elsarticle}
\graphicspath{ {./figures/} }
\usepackage{hyperref}
\usepackage{float}
\usepackage{verbatim} %comments
\usepackage{apalike}
\usepackage{multirow}
\usepackage{graphicx}
\restylefloat{figure}
\restylefloat{table}

\journal{Expert Systems with Applications}

\biboptions{authoryear}

\begin{document}
\begin{frontmatter}

\begin{titlepage}
\begin{center}
\vspace*{1cm}

\textbf{ \large PETA: Evaluating the Impact of Protein Transfer Learning with Sub-word Tokenization on Downstream Applications }

\vspace{1.5cm}

% Author names and affiliations
Yang Tan†$^{a,b}$ (tyang@mail.ecust.edu.cn), Mingchen Li†$^{a, b}$ (lmc@mail.ecust.edu.cn), Pan Tan$^{b, c}$ (tanpan@pjlab.org.cn), Ziyi Zhou$^{c}$ (zy-zhou@sjtu.edu.cn), Huiqun Yu*$^{a}$ (yhq@ecust.edu.cn), Guisheng Fan*$^{a}$ (gsfan@ecust.edu.cn), Liang Hong*$^{b, c}$ (hongl3liang@sjtu.edu.cn)  \\

\hspace{10pt}

\begin{flushleft}
\small  
$^a$ School of Information Science and Engineering, East China University of Science and Technology, Shanghai 200240, China \\
$^b$ Shanghai Artificial Intelligence Laboratory, Shanghai 200240, China \\
$^c$ Shanghai National Center for Applied Mathematics (SJTU Center), \& Institute of Natural Sciences, Shanghai Jiao Tong University

\begin{comment}
Clearly indicate who will handle correspondence at all stages of refereeing and publication, also post-publication. Ensure that phone numbers (with country and area code) are provided in addition to the e-mail address and the complete postal address. Contact details must be kept up to date by the corresponding author.
\end{comment}

\vspace{1cm}
\textbf{Corresponding Author:} \\
Huiqun Yu \\
School of Information Science and Engineering, East China University of Science and Technology, Shanghai 200240, China \\
Email: yhq@ecust.edu.cn

\textbf{Corresponding Author:} \\
Guisheng Fan \\
School of Information Science and Engineering, East China University of Science and Technology, Shanghai 200240, China \\
Email: gsfan@ecust.edu.cn

\textbf{Corresponding Author:} \\
Liang Hong \\
Shanghai Artificial Intelligence Laboratory, Shanghai 200240, China \\
Shanghai National Center for Applied Mathematics (SJTU Center), \& Institute of Natural Sciences, Shanghai Jiao Tong University \\
Email: hongl3liang@sjtu.edu.cn

\end{flushleft}        
\end{center}
\end{titlepage}

\title{PETA: Evaluating the Impact of Protein Transfer Learning with Sub-word Tokenization on Downstream Applications}

\author[label1,label2]{Yang Tan†}
\ead{tyang@mail.ecust.edu.cn}

\author[label1,label2]{Mingchen Li†\footnote{†: Yang Tan and Mingchen Li are co-first authors.}}
\ead{lmc@mail.ecust.edu.cn}

\author[label2,label3]{Pan Tan}
\ead{tanpan@pjlab.org.cn}

\author[label3]{Ziyi Zhou}
\ead{zy-zhou@sjtu.edu.cn}

\author[label1]{Huiqun Yu\corref{cor1} }
\ead{yhq@ecust.edu.cn}

\author[label1]{Guisheng Fan\corref{cor1}}
\ead{gsfan@ecust.edu.cn}

\author[label2,label3]{Liang Hong\corref{cor1}}
\ead{hongl3liang@sjtu.edu.cn}

\cortext[cor1]{Corresponding author.}
\address[label1]{School of Information Science and Engineering, East China University of Science and Technology, Shanghai 200240, China}
\address[label2]{Shanghai Artificial Intelligence Laboratory, Shanghai 200240, China}
\address[label3]{Shanghai National Center for Applied Mathematics (SJTU Center), \& Institute of Natural Sciences, Shanghai Jiao Tong University}

\begin{abstract}
Large protein language models are adept at capturing the underlying evolutionary information in primary structures, offering significant practical value for protein engineering. Compared to natural language models, protein amino acid sequences have a smaller data volume and a limited combinatorial space. Choosing an appropriate vocabulary size to optimize the pre-trained model is a pivotal issue. Moreover, despite the wealth of benchmarks and studies in the natural language community, there remains a lack of a comprehensive benchmark for systematically evaluating protein language model quality. Given these challenges, PETA trained language models with 14 different vocabulary sizes under three tokenization methods. It conducted thousands of tests on 33 diverse downstream datasets to assess the models' transfer learning capabilities, incorporating two classification heads and three random seeds to mitigate potential biases. Extensive experiments indicate that vocabulary sizes between 50 and 200 optimize the model, whereas sizes exceeding 800 detrimentally affect the model's representational performance. Our code, model weights and datasets are available at \url{https://github.com/ginnm/ProteinPretraining}.
\end{abstract}


\end{frontmatter}

\section{Introduction}
\label{introduction}

Naturally occurring proteins play a pivotal role in sustaining life forms and have found extensive applications in human endeavors, including gene editing \citep{doudna2014cas9, hsu2014cas9-2}, drug discovery \citep{scott2016drug}, and enzymatic catalysis \citep{lee2006enzymatic_catalysis}. Furthermore, gaining insights into protein properties or enhancing their functionality holds significant practical value, such as enhancing the function of the original protein \citep{joo1999enhance_function} or annotating an unknown sequence \citep{sharan2007anotate_protein}. Protein engineering typically follows two common approaches: laboratory-based experiments and computation-based methods. The former includes structural analysis \citep{feng2014structual_analysis}, expression purification \citep{lesley2001protein_expression} and direct evolution \citep{arnold1998direct}, while valuable, are time-consuming and heavily reliant on domain-specific knowledge. This limitation falls short of meeting the evolving demands of both the scientific community and industry. Conversely, the computation-based modeling strategy relies on machine-learning or physics-based methods that are often not particularly accurate but are cost-effective and time-saving. Thanks to the advancements in protein sequencing technology \citep{ma2015sequencing}, new avenues have opened up for training large-scale protein models capable of capturing a more comprehensive understanding. For instance, Meta has introduced ESM series \citep{lin2023esm2, meier2021esm_1v, rives2021esm_1b} to leverage the UniProt database \citep{uniprot2019uniprot} which contains over 200 million protein sequences or its subsets for training purposes.

A plethora of machine-learning methods have been developed to address various protein-related tasks. Protein representation learning primarily employs two main strategies. The first strategy, recognizing the significance of protein structures in functional determination, employs structure-aware encoders trained with implicit topological constraints. Notable examples include ESM-GearNet \citep{zhang2023gearnet} and LM-GVP \citep{wang2022lm_gvp}. However, residue-level structural models often exhibit inferior performance compared to protein language models (PLMs) across most tasks \citep{zhang2022structure_performance}. These methods, in contrast to PLMs, tend to focus on a limited and biased subset of the Protein Data Bank (PDB) \citep{berman2000pdb}, limiting their training data scope and generalization capabilities. While efforts have been made to expand the training dataset scale, such as incorporating AlphaFold database \citep{varadi2022afdb} data in some work \citep{hsu2022esm_if, zhang2023gearnet}, their parameters and data sizes still remain relatively small in comparison to language models.

The second strategy involves protein language processing, a well-established approach that models the inherent connections within protein sequences and effectively captures co-evolutionary information. Various works have harnessed different sequence encoders or decoders in this strategy, including Convolutional Neural Networks (CNNs) \citep{yang2022carp}, Long Short-Term Memory networks (LSTMs) \citep{alley2019unirep, rao2019tape}, BERT models \citep{lin2023esm2, meier2021esm_1v, rives2021esm_1b}, Generative Pre-trained Transformers (GPTs) \citep{madani2023progen, nijkamp2022progen2, ferruz2022protgpt2}, and Transformers \citep{elnaggar2021prottrans, elnaggar2023ankh}. Protein language models have demonstrated outstanding performance across a wide range of tasks and constitute a primary focus of this paper.

\begin{figure*}[!t]
    \centering
    \includegraphics[width=\textwidth]{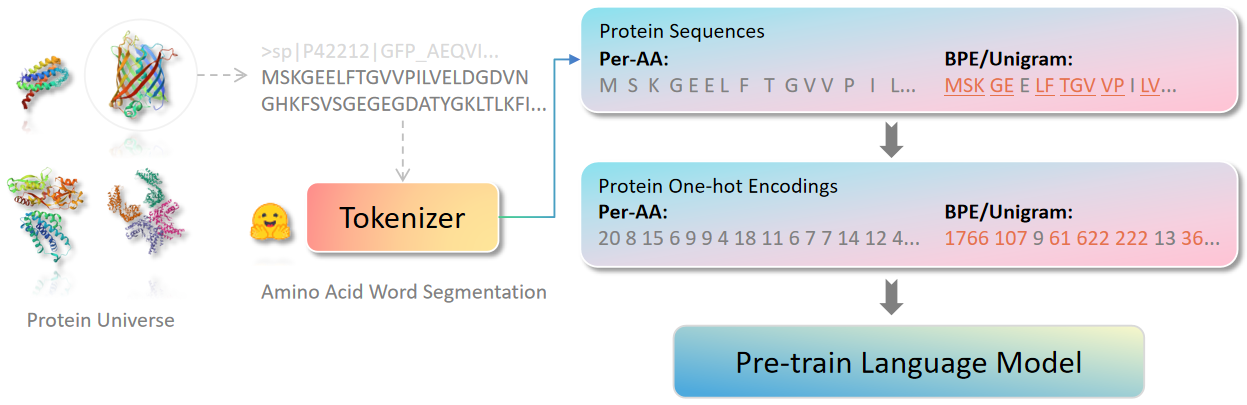}
    \caption{The protein sequence is formed into a new discrete token sequence through different word segmentation methods. As the size of the vocabulary increases, the amino acid composition of a single token becomes more complex.}
    \label{fig:tokenizer}
\end{figure*}

Nonetheless, there is still a lack of comprehensive analysis of how amino acid combinations influence language models, it's a fundamental and significant issue in language model training \citep{rust2020toeknizer_1, choo2023tokenizer_2} that has thus far been overlooked in the protein area. In this paper, we draw inspiration from \cite{asgari2019vocab} and aim to develop a universal amino acid coding approach capable of delivering robust performance across various protein-related tasks, while harnessing the benefits of knowledge sharing and transfer as shown in Fig \ref{fig:tokenizer}. To facilitate a thorough assessment and take cues from the success of benchmark datasets in domains like computer vision and natural language processing, e.g., ImageNet \citep{deng2009imagenet} and GLUE \citep{wang2018glue}, we have meticulously curated a collection of 33 datasets categorized into 15 distinct tasks. These datasets are integral to advancing the realm of deep learning in protein comprehension. Our PETA benchmark encompasses five groups, covering tasks in protein fitness prediction, protein localization prediction, protein-protein interaction prediction, protein solubility prediction, and protein structure prediction. For each individual dataset, we evaluate the performance of three types of tokenizers, two new residue-pair tokenizers are used to train five models with distinct vocabularies respectively and one per-amino-acid (Per-AA) model acts as baseline. Two different pooling mechanisms and three random seeds are employed in downstream tasks to mitigate potential classification biases. We anticipate that our comprehensive analysis of protein tokenizers and the PETA benchmark will serve as a pivotal milestone for the continued advancement of protein language model pre-training.

Our contributions are as follows:
\begin{itemize}
    \item \textbf{Protein Tokenization Analysis}: We summarize how the amino acid coding approach enhances the effectiveness of protein language models across diverse protein-related tasks. By addressing the influence of amino acid combinations, the research offers valuable insights into the optimization of protein language models.
    \item \textbf{Creation of the PETA Benchmark}: We meticulously curate the PETA benchmark, a comprehensive collection of 33 datasets categorized into 15 distinct protein-related tasks. This benchmark spans 5 diverse aspects of protein research. It provides a standardized evaluation framework for protein language models.
    \item \textbf{Comprehensive Experiments}: We have conducted thousands of experimental evaluations to ensure the validity of the results, and the model weights, code, etc. are completely open source in the community.
\end{itemize}

\section{Related work}
\label{title_page}

\subsection{Protein Representation Learning} 
Representation learning harnesses knowledge acquired from large-scale corpora to generalize across various tasks. Early approaches primarily employed machine learning techniques from natural language processing, such as word2vec \citep{mikolov2013word2vec} and doc2vec \citep{le2014doc2vec}, to extract features from protein sequences \citep{wang2017possum, mejia2019kmer, wan2016feature}. Recently, deep learning has exhibited tremendous potential by enabling models with increased capacity and deeper encoders, capable of handling millions or billions of protein sequences. ESM-1V \citep{meier2021esm_1v}, SESNet \citep{li2023sesnet}, and ECNet \citep{luo2021ecnet}, which focus on predicting mutation fitness. Additionally, Meta's ESM-1b \citep{rives2021esm_1b} and ESM-2 \citep{lin2023esm2} employ mask language modeling. ProtTrans \citep{elnaggar2021prottrans} pre-trains language models under various architectures \citep{lan2019albert, devlin2018bert, vaswani2017attention, raffel2020t5, ye2020xlnet}, while XTrimo \citep{chen2023xtrimopglm} aligns its pre-trained architecture with GLM \citep{du2021glm}. Ankn \citep{elnaggar2023ankh} uses an asymmetric encoder and decoder framework and different mask probabilities to improve the pre-training performance. CPCProt \citep{lu2020cpcprot} leverages a contrastive predictive coding loss, whereas ProGen \citep{madani2023progen, nijkamp2022progen2}, UniRep \citep{alley2019unirep}, ProXLNet \citep{elnaggar2021prottrans}, ProtGPT2 \citep{ferruz2022protgpt2}, and Tranception \citep{notin2022tranception} are pre-trained using next amino acid prediction tasks. Although many of these approaches share common objectives with natural language processing, there are also innovations like CARP \citep{yang2022carp} which employ convolutional networks for downstream tasks. Some works delve into protein multiple sequence alignments (MSAs) \citep{rao2021msa_transformer, biswas2021low_n, notin2022tranception}, while others take structure-based approaches to extract topology information for inverse folding \citep{hsu2022esm_if, yang2022mif, jing2020gvp} or protein design \citep{dauparas2022proteinmpnn, zheng2023lm_design}. Notably, LM-GVP \citep{wang2022lm_gvp} and MIF-ST \citep{yang2022mif} integrate sequence and structural information to enhance the learning of effective protein representations. In this benchmark, our primary focus revolves around evaluating the performance of language models utilizing different tokenization strategies.

\subsection{Protein Modeling Benchmarks}
A comprehensive benchmark has shown great influence in the traditional computer science community and driven the research direction of different works \citep{wang2018glue, wang2019superglue, lin2014coco, deng2009imagenet, kay2017kinetics}. However, it is worth noting that the field of computing protein engineering still lacks a large-scale benchmarking framework. In contrast, the biennial Critical Assessment of Protein Structure Prediction (CASP) \citep{kryshtafovych2021casp} has emerged as a gold standard for assessing advancements in protein structure prediction. In tandem with CASP, the Critical Assessment of Functional Annotation (CAFA) challenge \citep{zhou2019cafa} has been established to evaluate the prediction of protein functions. Several notable works, such as DeepSequence \citep{riesselman2018deepsequence}, Envision \citep{gray2018envision}, and ProteinGym \citep{notin2022tranception}, focus on measuring very different functional fitness variations in response to diverse protein modifications, including substitutions and insertions/deletions. Techniques like deep mutational scanning (DMS) \citep{fowler2014dms} and other protein engineering methods are used to build up these datasets. On the other hand, works like SoluProtMutDB \citep{velecky2022soluprotmutdb}, SKEMPI \citep{moal2012skempi}, and ProThermDB \citep{nikam2021prothermdb} concentrate on assessing specific properties concerning single amino acid variations (SAVs). Additionally, FLIP \citep{dallago2021flip} offers various data partitioning methods across three protein landscapes for evaluating fitness prediction. The TAPE benchmark \citep{rao2019tape} encompasses five tasks, with three focusing on structure prediction and the remaining two targeting fitness prediction. PEER \citep{xu2022peer} encompasses seventeen biologically relevant tasks spanning five aspects of protein understanding. ProteinGLUE \citep{capel2022proteinglue} comprises seven downstream tasks designed for self-supervised protein representation learning. DeepLoc \citep{almagro2017deeploc-1, thumuluri2022deeploc-2} provides datasets for subcellular localization classification. The STRING database \citep{szklarczyk2016string} annotates protein-protein interactions (PPIs) with seven types of interactions. TDA \citep{huang2021tda} generates protein-related datasets and tasks tailored for drug discovery. ESOL website \citep{niwa2009esol_dataset} aggregates solubility scores for ensemble E.coli proteins.

\section{PTEA}

\begin{figure*}[!t]
    \centering
    \includegraphics[width=\textwidth]{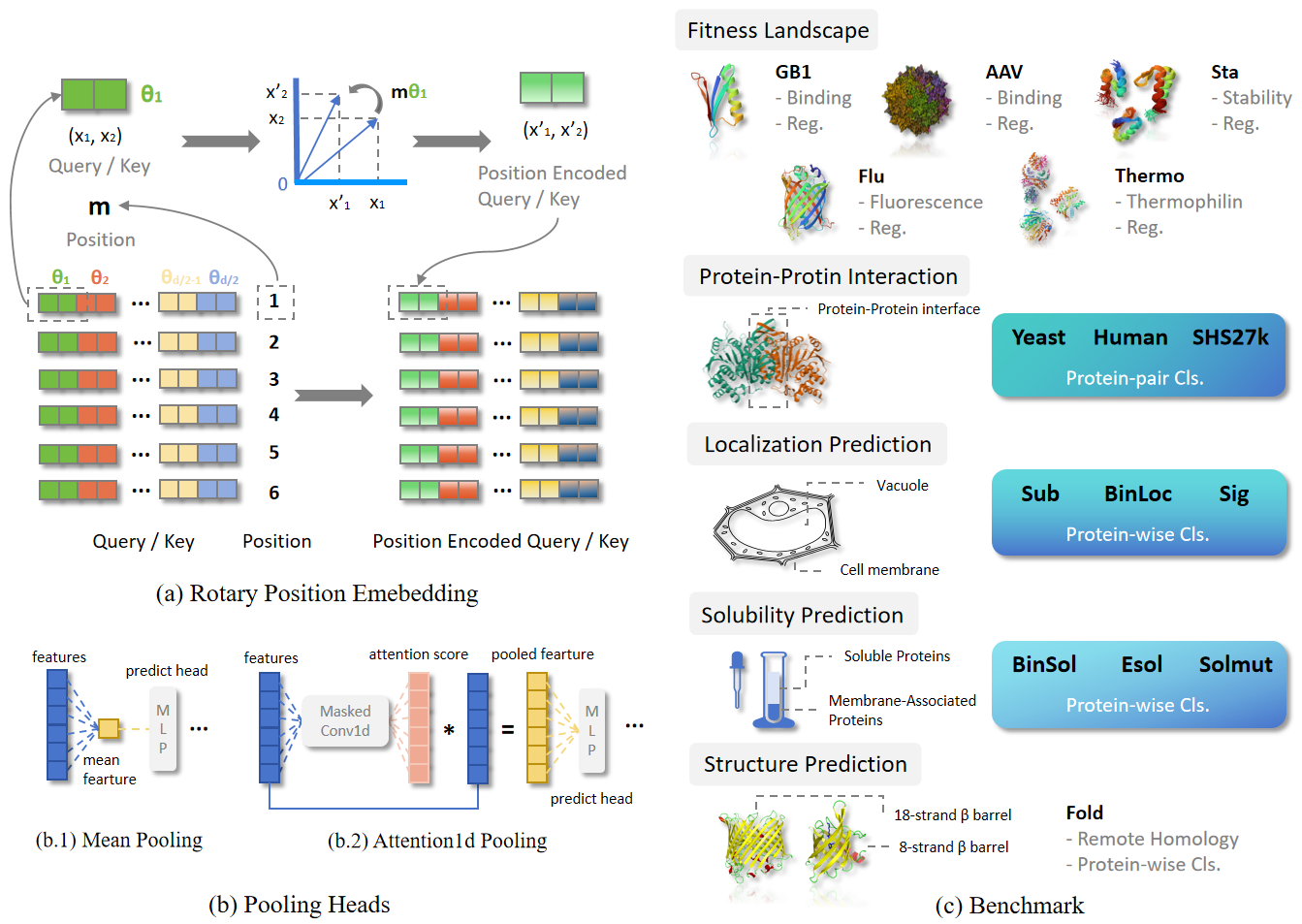}
    \caption{The framework of PETA. (a) Pre-trained models use rotary position embedding, which possesses favorable theoretical properties and is an absolute positional encoding applicable to linear Attention. (b) We employed two distinct classification heads, namely mean pooling and attention1d pooling. The former is the most commonly used method at present, while the latter is relatively more complex. (c) Our benchmark comprises 15 downstream tasks, which can be categorized into five groups. Some of these downstream tasks include multiple datasets or data splitting methods, amounting to a total of 33 datasets.}
    \label{fig:framework}
\end{figure*}

We designed PETA to answer two important questions:
\begin{itemize}
    \item Is residue-wise tokenizer good enough for protein language model pre-training?
    \item How do different vocab sizes influence the representation ability on downstream tasks?
\end{itemize}

Most of the works choose one tokenizer aligned with previous research without much concern, to answer the first question, we utilize three amino acid segmentation strategies including residue-wise and sub-word tokenizer. For the second question, we design vocab size from \{50, 100, 200, 800, 1600, 3200\} for the Unigram and BPE segmentation method. The model trained under per-AA is the baseline, it has a vocabulary size of $33$ and many works adopt this. In general, we utilize three tokenize methods, two types of classification heads and two model pipelines to solve different tasks in the PETA benchmark. The framework of PETA is shown in Fig \ref{fig:framework}.

\subsection{Amino Acid Segmentation}

In this study, we utilize three classic protein sequence segmentation methods: per-amino-acid encoding (Per-AA), byte pair encoding (BPE) \citep{gage1994bpe}, and unigram language modeling (Unigram) \citep{bengio2000unigram} as shown in Fig \ref{fig:tokenizer}. Per-AA focuses on individual amino acid units, enabling high-resolution analysis of subtle variations. BPE offers flexibility by segmenting sequences into subunits, effectively capturing structural information, while Unigram, based on character-level statistics, captures global sequence characteristics. These diverse methods collectively enhance our comprehensive analysis of protein sequences, each serving a unique role in addressing specific analytical requirements.

\subsection{Pre-training Protein Language Models}

\subsubsection{Model Architecture}
Our pre-training architecture employs RoFormer \citep{su2021roformer}, an autoencoding model that adopts a BERT-like structure augmented with rotary positional embeddings, as illustrated in Figure \ref{fig:framework} (a). These rotary positional embeddings effectively harness positional information within sequences. Detailed hyperparameter configurations are delineated in the subsequent section. Initially, protein sequences are tokenized and transformed into one-hot encoded representations. These representations are subsequently fed into RoFormer's encoder, which generates sets of hidden states that maintain the length consistency with the input tokenized sequence. Finally, these hidden states are transformed into a vector with a dimensionality corresponding to the vocabulary size, upon which a softmax function is applied to yield the reconstruction probability density distribution.

\subsubsection{Pre-training Objective}

We employ the masked language modeling (MLM) objective for pre-training our models \citep{lan2019albert}. Given an input sequence, a subset of tokens is selected at random and replaced with a special mask token. The model is then trained to predict these masked tokens based on the unmasked context tokens. The loss function  for this objective can be defined as:
\begin{equation}
    L_{MLM}=E_{x \sim X}E_{M}\sum_{x}-\log p(x_{i}|x_{/M})
\end{equation}

Here, $x$ is a sequence from the dataset $X$, and $x_{/M}$ represents the sequence with masked tokens removed. $p(x_{i}|x_{/M})$ is the conditional probability of predicting the correct token $x_i$ given the context $x_{/M}$. The aim is to minimize the negative log-likelihood of the true token at each masked index $i$, which in this case are amino acids, given the unmasked sequence as context. Intuitively, the model must learn to identify the dependencies between the masked and unmasked tokens to successfully predict the masked positions.

\subsubsection{Language Modeling Evaluation Metrics}
We use \textit{Perplexity} to evaluate the performance of the masked language modeling, computed as:
\begin{equation}
    Perplexity=\exp(-\frac1N\sum_{i=1}^N\log p(x_i|x))
\end{equation}
, where $N$ is the number of validation sequences, as well as $x_i$ is the $i_{th}$ sequence of validation set.
To account for potential unfair comparisons arising due to varying vocabulary sizes across different models, we introduce the metric of \textit{Normalized Perplexity}. The formula for Normalized Perplexity is as follows:
\begin{equation}
    Perplexity=\exp{(-\frac1{N\times V}\sum_{i=1}^N\log{p(x_i|x)})}
\end{equation}
, where $V$ is the vocab size.

\subsubsection{Pre-training Data}
We train all models on UniRef90 \citep{suzek2015uniref}, a comprehensive protein sequence database that contains approximately 138 billion sequences from diverse life forms. This large-scale database serves as a robust training set for capturing the underlying patterns and intricacies associated with protein sequences. For model validation, a subset of 100,000 sequences is reserved from the UniRef90 database. This validation set is instrumental in evaluating the generalization capability of the language models on unseen data. This setup ensures that the trained models are subjected to a variety of sequence patterns, thereby facilitating a more robust understanding of protein sequences. By reserving a significant number of sequences for validation, we also ensure an unbiased assessment of the model performance.

\subsection{Classification Head}

To avoid the potential bias caused by different classification methods, the \textit{Mean Pooling} and \textit{Attention1d Pooling} are adopted under our evaluation, as shown in Fig \ref{fig:framework} (b). The former is trained on the average of features aligned with the first dimension, MLP and Relu activation are used to make a prediction. The latter is trained on an attention machine and 1D convolution layer to map different weights to embed and predict the label.

\subsection{Model Pipeline}

\textbf{Protein-wise tasks.} Leaning a function $y=f_{\theta}(x)$ that maps protein $x$ to the label $y$, where $f_\theta$ is parameterized by a sequence-based encoder and a classification head defined upon the residue-wise or residue-pair protein embedding.

\textbf{Protein-pair tasks.} Leaning a function $y=f_{\theta}(x)$ that maps a pair of proteins ($x, x`$) to the label $y$, where $f_\theta$ is parameterized by a pair of siamese sequence-based encoders and a classifier defined upon the sum of the embeddings of two proteins.

\begin{table}[]
\caption{Bechmark task details. Each task, along with its task name, category, the count of datasets, the source of dataset and evaluation metric are shown below. \textit{Abbr.}, Reg.: regression; Cls.: classification; MSE: mean square error.}
\label{tab:benchamrk}
\resizebox{\textwidth}{!}{%
\begin{tabular}{ccccc}
\hline
Task Name                           & Task Category     & Count & Source & Metric   \\ \hline
\multicolumn{5}{c}{Fitness Prediction}                                                    \\ \hline
GB1 fitness prediction (GB1)        & Protein-wise Reg. & 5     & FLIP         & Spearman \\
AAV fitness prediction (AAV)        & Protein-wise Reg. & 7     & FLIP         & Spearman \\
Thermostability prediction (Thermo) & Protein-wise Reg. & 3     & FLIP         & Spearman \\
Fluorescence prediction (Flu)       & Protein-wise Reg. & 1     & TAPE         & Spearman \\
Stability prediction (Sta)          & Protein-wise Reg. & 1     & TAPE         & Spearman \\ \hline
\multicolumn{5}{c}{Protein-Protein Interaction Prediction}                                \\ \hline
Yeast PPI prediction (Yeast)        & Protein-pair Cls. & 1     & PETA         & Accuracy \\
Human PPI prediction (Human)        & Protein-pair Cls. & 1     & PETA         & Accuracy \\
SHS PPI prediction (SHS27k)         & Protein-pair Cls. & 1     & PETA         & Accuracy \\ \hline
\multicolumn{5}{c}{Localization Prediction}                                               \\ \hline
Subcellular localization prediction (Sub) & Protein-wise Cls. & 3 & pro-loc, DeepLoc-2 & Accuracy \\
Binary localization prediction (BinLoc)   & Protein-wise Cls. & 1 & pro-loc            & Accuracy \\
Sorting signal prediction (Sig)     & Protein-wise Cls. & 1     & DeepLoc-2    & Accuracy \\ \hline
\multicolumn{5}{c}{Solubility Prediction}                                                 \\ \hline
Binary solubility prediction (BinSol)     & Protein-wise Cls. & 1 & DeepSol            & Accuracy \\
E.coli solubility prediction (Esol) & Protein-wise Reg. & 1     & GraphSol     & MSE      \\
Mutation solubility prediction (Solmut)   & Protein-wise Reg. & 3 & PETA               & Spearman \\ \hline
\multicolumn{5}{c}{Structure Prediction}                                                  \\ \hline
Fold Prediction (Fold)              & Protein-wise Cls. & 3     & TAPE         & Accuracy \\ \hline
\end{tabular}%
}
\end{table}
\section{Benchmark Tasks}

The PETA benchmark includes 15 tasks within 5 groups and 33 datasets in total, mainly focusing on protein-wise and protein-pair tasks. We have curated tasks from influential protein engineering applications and made updates to certain datasets to ensure their relevance and accuracy, a summary of the downstream dataset statistics is shown in Table \ref{tab:benchamrk}.

\subsection{Fitness Prediction}

This set of tasks aims to forecast functional attributes of proteins, which may be either discrete or continuous in nature. 

\textbf{GB1 fitness prediction} assesses fitness scores among mutations within the GB1 landscape from \cite{wu2016gb1}. Given a protein sequence $x$, we map it to a regression value $y \in \mathrm{R}$, where a fitness score of 1 represents the wild-type and 0 indicates non-binding affinity. In our analysis, we utilize all the dataset splits proposed in FLIP \citep{dallago2021flip}, encompassing "one-vs-rest", "two-vs-rest", "three-vs-rest", "low-vs-high" and "Sampled".

\textit{Impact}: GB1 serves as the binding domain of protein G \citep{mccallister2000gb1_protein_G}, an immunoglobulin binding protein found in Streptococcal bacteria \citep{sauer1995gb1_streptococcal}. This task stands as a gold standard for investigating epistatic interactions.

\textbf{AAV fitness prediction} entails the evaluation of fitness values associated with Adeno-associated virus (AAV) capsid proteins \citep{girod2002vp1}. Given a protein sequence $x$, we establish a mapping to a regression value $y \in \mathrm{R}$, focusing on the mutational window spanning positions 561 to 588 from \cite{bryant2021aav}. We adopt all the dataset splits from FLIP \citep{dallago2021flip}, which includes "Mut-Des", "Des-Mut", "one-vs-rest", "two-vs-rest", "seven-vs-rest", "low-vs-high" and "Sampled".

\textit{Impact}: AAV proteins are responsible for facilitating the integration of a DNA payload into a target cell by the virus \citep{vandenberghe2009aav_therapy}. This task specifically addresses the prediction of fitness for an extended sequence subjected to mutations at select positions.

\textbf{Thermostability prediction} involves the analysis of protein melting curves, which are acquired through a mass spectrometry-based assay and meticulously sourced from \cite{jarzab2020meltome}. In this endeavor, we focus on a protein sequence $x$, which is drawn from a vast pool of 48,000 proteins spanning 13 diverse species. Our objective is to predict a thermostability score $y \in \mathrm{R}$. For this analysis, we have employed the dataset splitting strategies "Human", "mixed\_split", and "Human\_cell" as outlined in FLIP \citep{dallago2021flip}.

\textit{Impact}: Thermostable proteins \citep{yeoman2010thermostable_impact, haki2003thermostable_impact2} demonstrate an ability to endure higher temperature conditions for extended periods or function at an accelerated rate. This task aligns closely with applications in protein engineering, particularly within industrial settings, where the enhanced stability of proteins can yield substantial benefits.

\textbf{Fluorescence prediction} primarily focuses on forecasting the fitness of mutants of the green fluorescent protein (GFP) \citep{labas2002gfp}, as documented by \cite{sarkisyan2016gfp_dataset}. In this context, we are presented with a GFP mutant sequence $x$ and aim to predict the corresponding fluorescence intensity $y \in \mathrm{R}$. We leverage the dataset and split methodology derived from TAPE \citep{rao2019tape}, which involves training the model on lower-order mutants and subsequently evaluating it on higher-order mutants.

\textit{Impact}: Green fluorescent protein can mark particular proteins in an organic structure by its green fluorescence \citep{willig2006gfp_mark}, this makes it easier for researchers to observe. This task bears significance in uncovering mutational patterns that either enhance or diminish such vital biological properties.

\textbf{Stability prediction} endeavors to assess the resilience of proteins within their natural environment. It involves taking a protein sequence, denoted as $x$, and predicting its corresponding experimental stability score, denoted as $y\in \mathrm{R}$. In this pursuit, we leverage the dataset curated from \citep{rocklin2017stab_dataset} and employ the partitioning method introduced in TAPE \citep{rao2019tape}. The training dataset comprises proteins sourced from four rounds of experimental design, while the test dataset encompasses proteins that are Hamming distance-1 neighbors of the top candidate proteins.

\textit{Impact}: The design of stable proteins in the face of mutations plays a pivotal role in the field of protein engineering \citep{shoichet1995stab_impact}. This work is instrumental in various applications, such as ensuring the effective delivery of drugs before they degrade.

\subsection{Protein-Protein interaction Prediction}

Predicting protein-protein interactions (PPI) is pivotal for deciphering the intricate molecular networks underpinning cellular functions and disease mechanisms, guiding targeted therapeutic interventions.

\textbf{Yeast PPI prediction} involves the prediction of whether two yeast proteins interact with each other. When presented with two proteins, denoted as $x_1$ and $x_2$ from yeast, the classifier assigns a binary label $y \in {0,1}$ to signify the presence or absence of interaction between them. To accomplish this task, we utilize the yeast PPI dataset sourced from \cite{guo2008ppi_yeast_dataset}. In this dataset, half of the instances represent positive cases, selected from the DIP\_20070219 database of interacting proteins \citep{salwinski2004ppi_yeast_dip}, with stringent criteria that exclude proteins with fewer than 50 amino acids or exhibiting $\geq$40\% sequence identity. The negative cases are generated by randomly pairing proteins that lack evidence of interaction, and these pairs are further filtered based on their sub-cellular locations.

\textit{Impact}: The yeast dataset serves as a widely recognized benchmark \citep{hashemifar2018ppi_yeast_impact_1, yu2008ppi_yeast_impact_2} for assessing model performance, and yeast PPI prediction substantially enhances our comprehension of cellular processes by unveiling intricate protein interactions and providing crucial insights into the functional roles of proteins within yeast cells.

\textbf{Human PPI prediction} predicts whether two human proteins interact or not. When provided with protein sequences $x_1$ and $x_2$ from humans, the predictor generates a binary label $y \in {0,1}$ to indicate the presence or absence of interaction between them. We adopt dataset from \cite{pan2010ppi_human_dataset}, comprising positive protein pairs obtained from the Human Protein Reference Database (HPRD) \citep{peri2003hprd} and negative protein pairs sourced from different cellular compartments with no documented interaction \citep{rhodes2005ppi_human_neg}. To ensure data quality, self-interactions and duplicate interactions were removed, resulting in the creation of two datasets, namely "AB" and "CD." The "AB" dataset encompasses the entire dataset, while the "CD" dataset comprises selected proteins with identities below 25\%. For evaluation, we exclusively employ the "AB" split strategy in this task.

\textit{Impact}: Human PPI prediction holds immense practical significance in clinical research. Notably, insights into protein interactions linked to diseases enhance our understanding of human disease mechanisms, paving the way for innovative therapeutic strategies \citep{rual2005ppi_human_impact_1, yu2011ppi_human_impact_2}.

\textbf{SHS PPI prediction} is to classify the type of interaction between a given protein pair. Given two protein sequences, $x_1$ and $x_2$, the model aims to predict a label $y$ where $y \in \{0,1...,6\}$. These interaction types encompass categories such as "reaction", "activation", "catalysis", among others. Our analysis utilizes a dataset derived from interaction pairs specific to Homo sapiens, sourced from the STRING database \citep{szklarczyk2016string}. We adopt preprocess strategies as recommended by \cite{chen2019ppi_shs} where the SHS dataset is divided into two subsets: "SHS27k" and "SHS148k". For computational efficiency, our study focuses solely on the "SHS27k" subset. The data is partitioned into training, validation, and test sets at a random split ratio of 8:1:1.

\textit{Impact}: Understanding and categorizing the precise interactions between protein pairs is pivotal in unraveling intricate cellular mechanisms and shedding light on complex biological pathways. This knowledge not only aids in defining drug efficacy through network-based "drug-disease proximity measures" \citep{guney2016ppi_shs_impact_1} but also plays a crucial role in interpreting the outcomes of genome-wide association screens \citep{hillenmeyer2016ppi_shs_impact_2}.

\subsection{Localization Prediction}

Identifying the localization or local-related biological mechanism of proteins within various cellular compartments is of paramount importance in the process of functional annotation.

\textbf{Subcellular localization prediction} aims to dig out the specific cellular location of a given natural protein. Given a protein sequence denoted as $x$, the model assigns it to multiple possible localizations $y \in {0, 1, ..., 9}$, which may include designations such as "Nucleus" and "Cytoplasm", among others. To accomplish this task, we utilize two datasets from DeepLoc-1 \citep{almagro2017deeploc-1} and DeepLoc-2 \citep{thumuluri2022deeploc-2}. For the DeepLoc-1 dataset, we apply the split methodology introduced by \cite{stark2021deeploc-1_split}. Regarding the DeepLoc-2 dataset, its original split strategy involves 5-fold cross-validation. In our implementation, we employ the first three partitions as training data, the fourth as validation data, and ultimately evaluate the model's performance on the SwissProt localization dataset and the human protein atlas \citep{thul2017hpa}.

\textit{Impact}: The subcellular localization of proteins plays a crucial role in deciphering the fundamental mechanisms of diseases linked to abnormal subcellular localization \citep{delmolino2001sub_impact_1, millar2009sub_impact_2}. Notably, some proteins are recognized for their ability to localize within multiple cellular compartments, underscoring the intricate and pertinent nature of this research domain.

\textbf{Binary localization prediction} constitutes a sub-problem of the aforementioned task. The model's responsibility is to decide whether a given protein $x$ should be categorized as "membrane-bound" or "soluble," denoted as $y \in {0,1}$. The datasets for training and testing are drawn from DeepLoc-1 \citep{almagro2017deeploc-1}, which includes an additional label system where "S" represents soluble, "M" corresponds to membrane-bound, and "U" signifies unknown localization. To train the model, we employ the same data partitioning method as introduced by \cite{stark2021deeploc-1_split}, while excluding data points labeled as "U".

\textit{Impact}: Predicting protein localization as either membrane-bound or soluble is vital for deciphering cellular functions, particularly in signal transduction and transport \citep{gimpelev2004bin_impact}. It plays a pivotal role in drug discovery, enabling the design of targeted therapies against membrane proteins. 

\textbf{Sorting signal prediction} elucidates the intricate process of subcellular localization by identifying biological mechanisms within sorting signal sequences that guide proteins to specific subcellular structures or organelles. When presented with a short sequence $x$, the model assigns it to one of nine classes denoted as $y \in {0,1,...,8}$, encompassing designations such as "Signal Peptide (SP)" and "Mitochondrial Transit Peptide (MT)", among others. This constitutes a multi-label classification task, and we employ the dataset sourced from DeepLoc-2 \citep{thumuluri2022deeploc-2}. As the original dataset lacks an official split strategy, we perform a random split with a train/validation/test ratio of 8:1:1.

\textit{Impact}: Protein sorting signals facilitate the precise intracellular localization of proteins, thereby sustaining cellular homeostasis and the integrity of subcellular compartments \citep{kanner2003sig_impact_1, nielsen2019sig_impact_2}. These signals typically entail interactions with partner proteins or sorting complexes, it is significant to investigate protein sorting signals for comprehending the intracellular localization and functional intricacies of proteins.

\subsection{Solubility Prediction}

This group of tasks is to predict the protein solubility, which is critical for optimizing protein expression, purification, and drug development processes.

\textbf{Binary solubility prediction} aims to determine whether a protein is soluble or insoluble. When presented with a protein denoted as $x$, the model assigns it to a binary label, $y \in {0,1}$. The dataset and data partitioning approach are drawn from DeepSol \citep{khurana2018deepsol}, where protein sequences exhibiting a sequence identity of $\geq$ 30\% to any sequence in the test set are excluded from the training set. This task shares similarities with binary localization prediction but explicitly focuses on modeling solubility.

\textit{Impact}: Protein solubility is pivotal for swiftly and efficiently selecting appropriate protein samples, saving resources and time, particularly in biotechnology, drug development, and laboratory protein purification \citep{davis1999binsol_impact_1, trainor2017binsol_impact_2}. It improves experiment success rates and resource allocation while advancing scientific research.

\textbf{E.coli solubility prediction} involves forecasting the solubility value of E. coli proteins using an ensemble database, downloadable from the eSOL website \citep{niwa2009esol_dataset}. When provided with a sequence from E. coli, the model predicts a regression value, denoted as $y \in \mathrm{R}$. Solubility, in this context, is defined as the ratio of the supernatant fraction to the total fraction, as determined in physiochemical experiments referred to as PURE \citep{shimizu2005pure}. We utilize the training and validation datasets sourced from GraphSol \citep{chen2021graphsol} and further partition the validation dataset into separate validation and test sets.

\textit{Impact}: E. coli, as a prevalent host organism for protein expression, demands precise solubility predictions to optimize recombinant protein production, purification, and subsequent functional studies \citep{costa2014esol_impact_1}. Such predictions, based on experimental data and computational models, facilitate the selection of suitable protein candidates for diverse applications \citep{hebditch2017esol_impact_2}, ranging from structural biology to drug discovery and industrial processes.

\textbf{Mutation solubility prediction} measures the impact of mutations on protein solubility. Given a mutated protein sequence denoted as $x$, the model predicts the solubility change $y \in \mathrm{R}$ relative to the wild-type sequence. This task encompasses three distinct protein mutation datasets, with mutations occurring at single points within proteins such as "Beta-lactamase TEM (blat)", "Chalcone Synthase (cs)" and "Levoglucosan Kinase (lgk)". These datasets were sourced from SoluProtMutDB \citep{velecky2022soluprotmutdb} which provides manually curated and reliable data in the standardized format. Data points where recorded mutations did not align with the original sequence were excluded, and the training, validation, and test datasets were partitioned in an 8:1:1 ratio.

\textit{Impact}: Low protein solubility is a significant hurdle in industrial processes and is implicated in numerous human diseases \citep{musil2018solmut_impact_1}. Investigating the impact of mutations on protein solubility not only sheds light on the mechanisms underpinning disease development but also enhances the application of protein engineering in various industrial domains.

\subsection{Structure Prediction}

While AlphaFold \citep{jumper2021alphafold} and RoseTTAFold \citep{baek2021rosettafold} have made significant strides in structure prediction, the structure prediction related task still is a rigorous assessment to evaluate the representation quality of the sequence model.

\textbf{Fold prediction} is the automated classification of protein sequences into one of 1,195 known protein folds, facilitating the modeling of the sequence-structure relationship. Given any sequence $x$, the objective is to predict the fold label $y \in {0,1,...,1194}$, determined by the backbone coordinates of the corresponding protein structure. This task utilizes the dataset from \cite{hou2018deepsf}, originally derived from the SCOP 1.75 database \citep{fox2014scope}. Notably, this dataset meticulously addresses homologous sequence redundancy between test and training datasets through two distinct strategies: a three-level redundancy reduction at fold/superfamily/family levels and sequence identity reduction.

\textit{Impact}: Fold prediction is essential for unraveling the intricate relationship between a protein's primary sequence and its three-dimensional structure, with profound implications for fields ranging from structural biology to drug design \citep{chen2016profold}.

\section{Experiments}

\subsection{Experimental Setups}

We perform the pre-training of our models on 8  A100-80GB GPUs, using a data-parallel distribution strategy. The global batch size is set to 1024 (local batch size is set to 32), and the maximum sequence length is constrained to 1024 tokens. We employ dropout regularization with a rate of 0.1 during the pre-training phase to mitigate overfitting. The architecture comprises 12 encoder layers, with each layer having a hidden size of 768 and an intermediate size of 3072. The multi-head attention mechanism contains 12 heads, each with a dimensionality of 64. For model optimization, we utilize the Adam optimizer, with a learning rate initialized at 1e-4. The maximum number of training steps is set to 530,000. The learning rate schedule involves a warm-up mechanism for the first 2000 iterations, following which the learning rate is linearly decayed to zero. The Adam hyperparameters are configured as follows: epsilon is 1e-8, $\beta_1=0.9$ and $\beta_2=0.98$. Gradient clipping is applied with a maximum value of 5.0 to prevent exploding gradients. Our implementation leverages the PyTorch framework in conjunction with the Hugging Face library, aligning with best practices for efficient and scalable training of language models.

In the case of supervised tasks, all pre-trained model weights are kept fixed to ensure a fair evaluation of their representation capabilities. Classifiers are trained using a batch size of 256, a learning rate of 0.001, and the Adam optimizer. Early stopping is employed with a patience threshold of 20 epochs, with a maximum of 100 epochs for training. It's important to note that these hyperparameters were adopted without adjustments, drawing reference from \cite{dallago2021flip}. Each individual experiment underwent training three times using different random seeds, and the final results represent the average scores obtained.

\begin{table}[]
\centering
\resizebox{\textwidth}{!}{%
\begin{tabular}{cccccccc}
\hline
Tokenization & \multicolumn{7}{c}{BPE} \\ \hline
Vocab size & 50 & 100 & 200 & 400 & 800 & 1600 & 3200 \\
Perplexity & 9.51 & 13.66 & 23.67 & 34.87 & 49.64 & 72.39 & 105.01 \\
Normalized Perplexity & 1.05 & 1.03 & 1.02 & 1.01 & 1.00 & 1.00 & 1.00 \\ \hline
\end{tabular}%
}
\caption{Perplexity and Normalized Perplexity on the validation set for the BPE model.}
\label{tab:bpe_pretrain}
\end{table}

\begin{table}[]
\centering
\resizebox{\textwidth}{!}{%
\begin{tabular}{cccccccc}
\hline
Tokenization & \multicolumn{7}{c}{Unigram} \\ \hline
Vocab size & 50 & 100 & 200 & 400 & 800 & 1600 & 3200 \\
Perplexity & 8.26 & 12.95 & 26.98 & 62.39 & 81.11 & 115.90 & 220.77 \\
Normalized Perplexity & 1.04 & 1.03 & 1.02 & 1.01 & 1.01 & 1.00 & 1.00 \\ \hline
\end{tabular}%
}
\caption{Perplexity and Normalized Perplexity on the validation set for the Unigram model.}
\label{tab:unigram_pretrain}
\end{table}

\subsection{Pre-training Results}
Following the pre-training phase, all models achieved a reduction in loss to an acceptable level, demonstrating effective learning from the training data. Table \ref{tab:bpe_pretrain} and Table \ref{tab:unigram_pretrain} present the perplexity and normalized perplexity metrics calculated on the test set for both Byte Pair Encoding (BPE) and Unigram models, respectively. For the base model employing a per-amino-acid (Per-AA) tokenization strategy, the \textit{Perplexity} value is $7.78$, and the corresponding \textit{Normalized Perplexity} is $1.06$.

In the supplementary materials, Figures S1 to S30 show the loss curves, as well as the perplexity and normalized perplexity curves for the pre-trained models. It is important to note that evaluations were performed at intervals of 10,000 steps. These figures collectively demonstrate that all models have converged to a reasonable range, substantiating their effectiveness in learning the underlying data distribution.

\subsection{Benchmark Results Overview}

\begin{table}[]
\centering
\resizebox{\textwidth}{!}{%
\begin{tabular}{cccccccccccccccc}
\hline
\multirow{2}{*}{Vocab} & \multicolumn{5}{c}{Fitness} & \multicolumn{3}{c}{PPI} & \multicolumn{3}{c}{Localization} & \multicolumn{3}{c}{Solubility} & Structure \\ \cline{2-16} 
 & GB1 & AAV & Thermo & Flu & Sta & Yeast & Human & SHS & Sub & BinLoc & Sig & BinSol & Esol & Solmut & Fold \\ \hline
50 & 4/5 & 2/7 & 2/3 & 1/1 & 1/1 & 1/1 & 1/1 & 1/1 & 3/3 & 1/1 & 1/1 & 1/1 & 0/1 & 3/3 & 0/3 \\
100 & 5/5 & 0/7 & 2/3 & 0/1 & 1/1 & 1/1 & 0/1 & 1/1 & 3/3 & 1/1 & 1/1 & 1/1 & 0/1 & 3/3 & 0/3 \\
200 & 4/5 & 2/7 & 2/3 & 1/1 & 0/1 & 1/1 & 1/1 & 1/1 & 3/3 & 1/1 & 1/1 & 1/1 & 0/1 & 3/3 & 0/3 \\
800 & 2/5 & 0/7 & 2/3 & 1/1 & 0/1 & 1/1 & 1/1 & 1/1 & 3/3 & 0/1 & 0/1 & 1/1 & 0/1 & 3/3 & 0/3 \\
1600 & 2/5 & 0/7 & 2/3 & 1/1 & 0/1 & 1/1 & 0/1 & 1/1 & 3/3 & 0/1 & 0/1 & 1/1 & 0/1 & 3/3 & 0/3 \\
3200 & 2/5 & 0/7 & 2/3 & 1/1 & 0/1 & 1/1 & 0/1 & 1/1 & 3/3 & 0/1 & 0/1 & 1/1 & 0/1 & 3/3 & 0/3 \\ 
\hline
\end{tabular}%
}
\caption{Average results on all experimental sets of each benchmark dataset. The denominator represents the number of data sets in this experiment, and the numerator represents the number of results that exceed the amino acid segmentation in this vocabulary size setting.}
\label{tab:result_overview}
\end{table}

To provide researchers with insights into how the augmentation of vocabulary size in Protein Language Models (PLMs) affects embedding quality, we conducted a systematic investigation. The scores in Table \ref{tab:result_overview} represent the average performance of models that utilize sub-word tokenization to expand their vocabulary. Specifically, for a given vocabulary size, it indicates how many datasets, on average, surpassed the traditional Per-AA method (with a vocabulary size of 33). This is calculated as the ratio of the number of datasets where the expanded vocabulary models outperformed the Per-AA baseline to the total number of datasets across various tasks. For models with expanded vocabularies, the average score on each dataset was computed from the mean of 12 experiment results (2 tokenization methods x 2 classification heads x 3 random seeds). For the baseline models, the average score on each dataset was derived from the mean of 6 experiments (2 classification heads x 3 random seeds). More detailed results of each experimental setting can be found in the supplementary materials Table S1 to S54.

Our experimental findings led to several key insights:
\begin{itemize}
\item \textbf{Significant Impact of Vocabulary Size.} Extensive experimentation has unequivocally demonstrated that vocabulary size profoundly influences protein representation, albeit with varying degrees of impact across different types of downstream tasks. Notably, in every dataset associated with structure prediction, an inverse relationship was observed wherein enhancements in vocabulary size correlated with negative optimization.
\item \textbf{Existence of an Optimal Vocabulary Threshold.} Contrasting with language models utilized in NLP, PLMs with an excessively large vocabulary size can potentially exert detrimental effects on downstream tasks. Specifically, when the vocabulary size surpasses 800, the majority of tasks are performed suboptimally compared to the baseline model that employs per-amino-acid segmentation.
\end{itemize}

\begin{table}[]
\centering
\resizebox{\textwidth}{!}{%
\begin{tabular}{ccccccccccc}
\hline
\multirow{2}{*}{Vocab} & \multicolumn{5}{c}{Fitness Prediction} & \multicolumn{5}{c}{Localization Prediction} \\ \cline{2-11} 
 & GB1* & AAV* & Thermo* & Flu & Sta & Sub-1 & BinLoc & Sub-2 & Sub-hpa & Sig \\ \hline
33 & 0.486 & \textbf{0.348} & 0.613 & 0.393 & 0.513 & \textbf{0.951} & 0.913 & \textbf{0.927} & 0.890 & 0.957 \\
50 & 0.503 & 0.318 & 0.618 & 0.399 & \textbf{0.530} & 0.948 & \textbf{0.917} & 0.926 & 0.891 & \textbf{0.961} \\
100 & \textbf{0.534} & 0.313 & \textbf{0.621} & 0.383 & 0.516 & 0.947 & 0.913 & 0.923 & 0.890 & 0.959 \\
200 & 0.504 & 0.306 & 0.616 & 0.420 & 0.495 & 0.945 & 0.913 & 0.922 & \textbf{0.894} & 0.959 \\
800 & 0.462 & 0.288 & 0.606 & 0.415 & 0.464 & 0.944 & 0.909 & 0.921 & \textbf{0.894} & 0.956 \\
1600 & 0.465 & 0.274 & 0.613 & \textbf{0.438} & 0.432 & 0.944 & 0.905 & 0.921 & 0.892 & 0.956 \\
3200 & 0.469 & 0.299 & 0.608 & 0.430 & 0.445 & 0.943 & 0.909 & 0.921 & 0.890 & 0.956 \\ \hline
\end{tabular}
}
\caption{Performance on \textbf{Fitness Prediction} and \textbf{Localization Prediction}. Each value indicates the average score across all experiments, maintaining a consistent vocabulary size. The averaging method is consistent with that in Section 5.2. Datasets marked with * indicate that averages were also taken across multiple data partitioning methods. For instance, \textbf{GB1} encompasses five different data segmentation methods within the same dataset. The score with a vocabulary size of 50 reflects results across 60 experimental setups (5$\times$2$\times$2$\times$3, representing the number of data splits, tokenization methods, classification heads, number of random seed experiments).}
\label{tab:task_set1}
\end{table}

\begin{figure*}[!t]
    \centering
    \includegraphics[width=\textwidth]{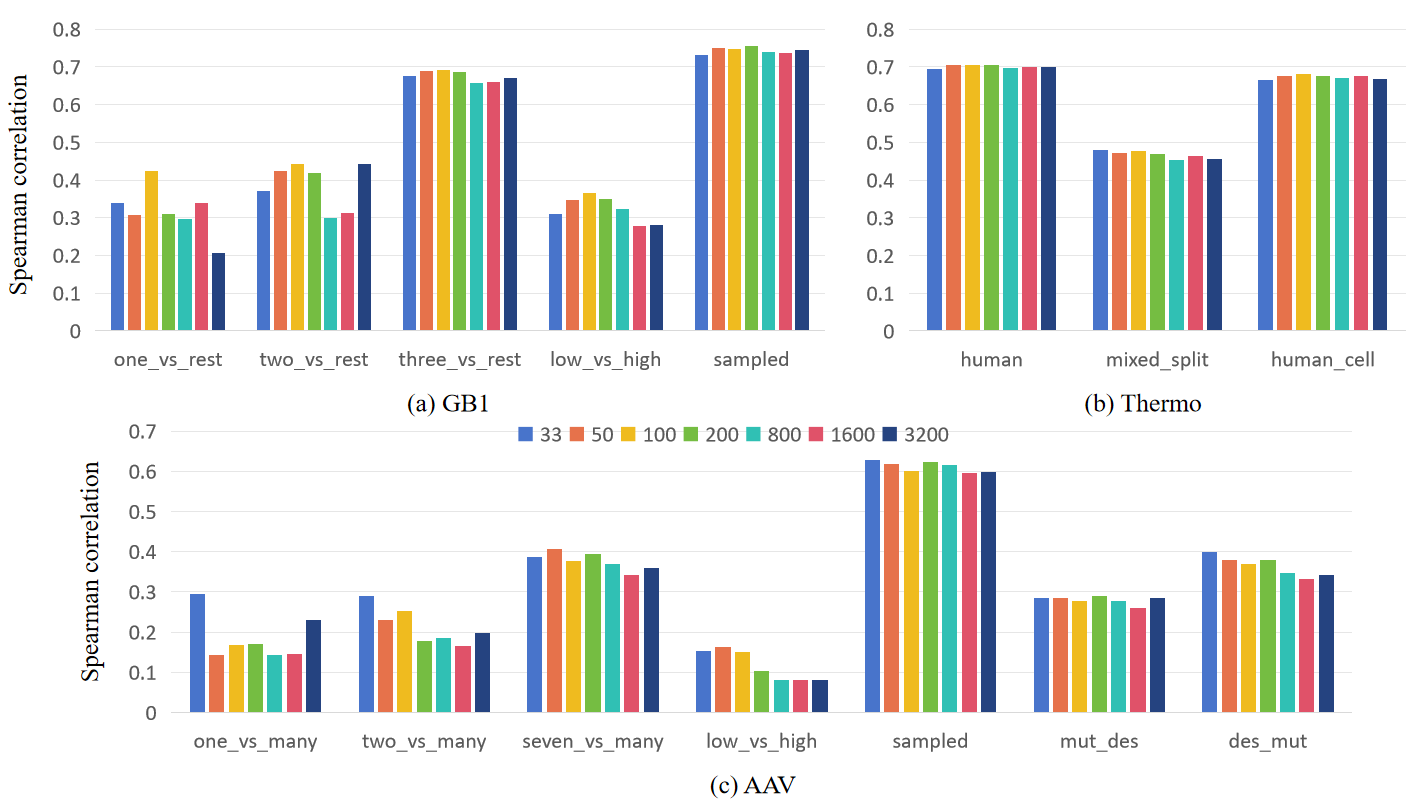}
    \caption{A detailed elaboration of the performance results for the \textbf{GB1}, \textbf{Thermo}, and \textbf{AAV} tasks across different vocabulary sizes is provided.}
    \label{fig:fitness}
\end{figure*}

\subsection{Downstream Tasks}

\textbf{Fitness Prediction.} Table \ref{tab:task_set1} showcases results for five distinct tasks under the umbrella of \textbf{Fitness Prediction} and the evaluation metrics is \textit{Spearman correlation}. It's worth highlighting that the datasets for \textbf{GB1}, \textbf{AAV}, and \textbf{Thermo} are multifaceted and are elaborated upon in Fig \ref{fig:fitness}. A discerning observation from the data is the dichotomous impact of vocabulary augmentation. Most tasks witnessed an upswing in their performance metrics with an expanded vocabulary. However, \textbf{AAV} stood out as an anomaly, showing a marked deterioration in performance across all its datasets as the vocabulary size expanded, the maximum decrease was 21\%, and the minimum decrease was 8.6\%. In contrast, the performance of \textbf{Flu} benefited from an enlarged vocabulary. Intriguingly, the average performance for \textbf{GB1} and \textbf{Stab} began to wane after hitting a vocabulary size of 200, even underperforming the baseline model set at a vocabulary size of 33, which notably lacked any word segmentation methodologies. \textbf{Thermo} is not sensitive to vocabulary size changes, fluctuating approximately 1\% up or down.

\textbf{Localization Prediction.} The right side of Table \ref{tab:task_set1} presents results from five datasets under the category of \textbf{Localization Prediction}, and the monitor metrics is \textit{Accuracy}. Across all datasets and partitioning methodologies, the task of subcellular localization prediction consistently achieves a remarkable classification accuracy exceeding 0.9. This accuracy remains predominantly stable, all performance fluctuations are within 1\% with variations in vocabulary size. Drawing from these experimental insights, it's evident that language models find both multi-class and single-class prediction tasks for protein localization relatively straightforward. Moreover, the vocabulary size seems to have minimal influence on the prediction outcomes for protein localization tasks.

\begin{table}[]
\caption{Performance on \textbf{PPI Prediction}, \textbf{Solubility Prediction} and \textbf{Structure Prediction}. Each value indicates the average score across all experiments, maintaining a consistent vocabulary size. The averaging method can be found in Section 5.2. Note that the \textit{Solmut} marked with * indicates that the average score was taken across multiple datasets.}
\resizebox{\textwidth}{!}{%
\begin{tabular}{cccccccccc}
\hline
\multirow{2}{*}{Vocab} & \multicolumn{3}{c}{PPI Prediction} & \multicolumn{3}{c}{Solubility Prediction} & \multicolumn{3}{c}{Structure Prediction} \\ \cline{2-10} 
     & Yeast          & SHS27k         & Human          & Esol           & BinSol         & Solmut*         & superfamily    & family         & fold           \\ \hline
33   & 0.699          & 0.519          & 0.927          & \textbf{0.044} & 0.913          & 0.187          & \textbf{0.518} & \textbf{0.930} & \textbf{0.269} \\
50   & 0.725          & 0.526          & \textbf{0.930} & 0.045          & \textbf{0.917} & \textbf{0.217} & 0.505          & 0.929          & 0.268          \\
100  & 0.720          & 0.520          & 0.925          & 0.048          & 0.913          & 0.208          & 0.437          & 0.919          & 0.226          \\
200  & 0.728          & \textbf{0.535} & 0.929          & 0.047          & 0.913          & 0.201          & 0.439          & 0.915          & 0.264          \\
800  & 0.720          & 0.534          & 0.928          & 0.046          & 0.909          & 0.210          & 0.435          & 0.894          & 0.240          \\
1600 & \textbf{0.733} & 0.531          & 0.925          & 0.048          & 0.905          & 0.213          & 0.418          & 0.890          & 0.231          \\
3200 & 0.724          & 0.520          & 0.922          & 0.047          & 0.909          & 0.197          & 0.406          & 0.880          & 0.233          \\ \hline
\end{tabular}%
}
\caption{Performance on \textbf{PPI Prediction}, \textbf{Solubility Prediction} and \textbf{Structure Prediction}. Each value indicates the average score across all experiments, maintaining a consistent vocabulary size. The averaging method is consistent with that in Section 5.2. Datasets marked with * indicate that averages were also taken across multiple data partitioning methods.}
\label{tab:task_set2}
\end{table}

\begin{figure*}[!t]
    \centering
    \includegraphics[width=\textwidth]{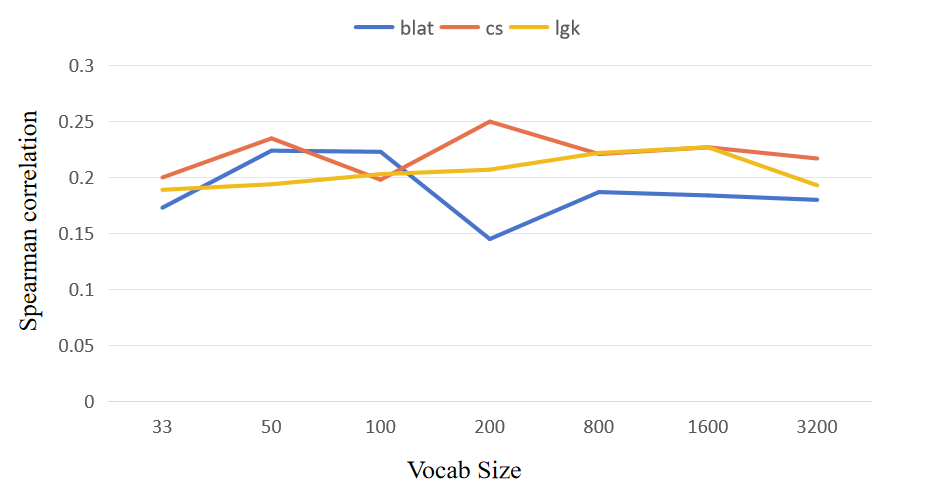}
    \caption{A detailed exposition is provided on the performance results of three distinct protein solubility mutation datasets: \textbf{Beta-lactamase TEM (blat)}, \textbf{Chalcone Synthase (cs)}, and \textbf{Levoglucosan Kinase (lgk)} across varying vocabulary sizes.}
    \label{fig:solmut}
\end{figure*}

\textbf{Protein-Protein interaction Prediction.} Table \ref{tab:task_set2} summaries the 3 datasets from \textbf{PPI Prediction}, and the metrics is \textit{Accuracy}. From the table, it's readily apparent that when dealing with protein pairing tasks, more discrete tokens are beneficial for distinguishing the relationships between protein pairs. Additionally, datasets that are harder to classify witness higher performance enhancements. For instance, in the \textbf{Yeast} dataset, the model with a vocabulary size of 1600 exhibited a 5\% average score increase compared to the model with a vocabulary size of 33. In \textbf{SHS27k}, every increase in vocabulary size resulted in performance improvements, with a peak boost of 3\%. In contrast, while the \textbf{Human} dataset showed improvements across the board, the maximum increase was a mere 0.4\%.

\textbf{Solubility Prediction.} The central section of Table \ref{tab:task_set2} displays the findings for \textbf{Solubility Prediction}. While the evaluation metric for the three datasets in Solmut is the \textit{Spearman correlation}, \textbf{Esol} employs \textit{MSE}, \textbf{BinSol} and \textbf{DeepSol} use \textit{Accuracy} as their metric. Predicting solubility regression values for \textbf{Esol} proves to be relatively straightforward, given the lower MSE scores and models of varying vocabulary sizes exhibit almost uniform predictive capabilities. \textbf{BinSol} observed a similar situation to \textbf{Esol}, the numerical fluctuations of results are less than 1\%. An analysis of the \textbf{Solmut} datasets indicates that models with expanded vocabulary sizes have the potential to improve performance by 20\% to 30\% as shown in Fig \ref{fig:solmut}. Although most instances show enhancement, occasional instability is detected. Such variability could be attributed to the alterations in the inherent characteristics of proteins post-mutation.

\textbf{Structure Prediction.} The results for \textbf{Structure Prediction} are detailed on the right side of Table \ref{tab:task_set2}, and the metrics is \textit{Accuracy}. In the context of fold prediction, a notable trend emerges: as the vocabulary size enlarges, there is a pronounced decrease in performance across all datasets. To illustrate, the \textbf{superfamily} category registers the most significant drop, plummeting by 21.6\%, while the \textbf{family} and \textbf{fold} categories experience declines of 5.3\% and 16\% respectively. This pattern underscores a consistent downturn in performance correlating with the augmentation of the vocabulary size. Such regression may stem from the amalgamation of multiple amino acid tokens in the encoding process, possibly veiling intricate local structural nuances. As a result, the surge in diversity might be detrimentally affecting the classification efficacy.

\subsection{Ablation Study}
\begin{table}[]
\caption{
The average results of different downstream task groups under the same vocabulary with varying tokenization methods are presented. Each score represents the average score of all experiments within that task group, encompassing different tasks, datasets, classification heads, and random seeds.}
\label{tab:ablation_tokenizer}
\resizebox{\textwidth}{!}{%
\begin{tabular}{ccccccccccc}
\hline
\multirow{2}{*}{Vocab} & \multicolumn{5}{c}{BPE}               & \multicolumn{5}{c}{Unigram}           \\ \cline{2-11} 
                            & Fit   & PPI   & Loc   & Sol   & Stru  & Fit   & PPI   & Loc   & Sol   & Stru  \\ \hline
50                          & 0.470 & 0.722 & 0.934 & 0.456 & 0.558 & 0.477 & 0.731 & 0.933 & 0.453 & 0.576 \\
100                         & 0.471 & 0.721 & 0.932 & 0.446 & 0.534 & 0.476 & 0.722 & 0.930 & 0.458 & 0.521 \\
200                         & 0.474 & 0.731 & 0.932 & 0.454 & 0.549 & 0.462 & 0.731 & 0.931 & 0.449 & 0.530 \\
800                         & 0.454 & 0.728 & 0.930 & 0.456 & 0.523 & 0.441 & 0.726 & 0.928 & 0.456 & 0.523 \\
1600                        & 0.452 & 0.732 & 0.929 & 0.457 & 0.515 & 0.437 & 0.727 & 0.927 & 0.459 & 0.510 \\
3200                        & 0.452 & 0.723 & 0.928 & 0.443 & 0.509 & 0.448 & 0.722 & 0.929 & 0.453 & 0.504 \\ \hline
\end{tabular}%
}
\end{table}

\begin{table}[]
\caption{The average results of different downstream task groups under the same vocabulary with varying pooling heads are presented. Each score represents the average score of all experiments within that task group, encompassing different tasks, datasets, classification heads, and random seeds.}
\label{tab:ablation_pool}
\resizebox{\textwidth}{!}{%
\begin{tabular}{ccccccccccc}
\hline
\multirow{2}{*}{Vocab} & \multicolumn{5}{c}{Mean Pooling}              & \multicolumn{5}{c}{Attention1d Pooling}       \\ \cline{2-11} 
                            & Fit   & PPI   & Loc   & Sol   & Stru  & Fit   & PPI   & Loc   & Sol   & Stru  \\ \hline
50                          & 0.411 & 0.706 & 0.930 & 0.418 & 0.562 & 0.537 & 0.748 & 0.937 & 0.490 & 0.572 \\
100                         & 0.415 & 0.701 & 0.924 & 0.420 & 0.526 & 0.531 & 0.742 & 0.938 & 0.484 & 0.529 \\
200                         & 0.412 & 0.708 & 0.925 & 0.427 & 0.543 & 0.524 & 0.753 & 0.937 & 0.476 & 0.536 \\
800                         & 0.393 & 0.700 & 0.922 & 0.429 & 0.520 & 0.501 & 0.755 & 0.936 & 0.483 & 0.526 \\
1600                        & 0.392 & 0.697 & 0.920 & 0.432 & 0.514 & 0.497 & 0.761 & 0.936 & 0.484 & 0.512 \\
3200                        & 0.398 & 0.687 & 0.921 & 0.415 & 0.506 & 0.502 & 0.757 & 0.936 & 0.481 & 0.507 \\ \hline
\end{tabular}%
}
\end{table}

\textbf{Analysis of tokenizers.}
From Table \ref{tab:ablation_tokenizer}, it can be observed that the discrepancies arising from different tokenization methods are minimal across various downstream tasks. The main source of performance variation stems from the impact of vocabulary size on model representation. Across different tasks, as the vocabulary size increases, the model performance exhibits a bell-shaped curve, showing an initial increase followed by a decline.

\textbf{Impact of pooling heads.}
From Table \ref{tab:ablation_pool}, it can be observed that, when freezing the pre-trained model parameters and only tuning the pooling head, the performance is highly correlated with the choice of the classification head. When using the same vocabulary size, the attention1d pooling method outperforms the mean pooling method. Additionally, similar to the results in Table \ref{tab:ablation_tokenizer}, as the vocabulary size increases, the model's representational capacity across various downstream tasks tends to decline.

\section{Conclusion}

In this paper, we introduce PETA, a vocabulary study optimized for protein language models across a broad range of datasets. To mitigate potential biases arising from different tokenization methods, classification heads, and random seeds, for each fixed vocabulary size, we employed both BPE and Unigram tokenization methods, two classification heads (mean pooling and attention1d pooling), and experiments with three different random seeds on each dataset. Ultimately, we found that expanding the vocabulary size to some extent (50-200) generally enhances performance on downstream tasks. However, once the vocabulary size surpasses 800, the model's representational power exhibits a broad decline across most tasks. We hope that this work and benchmark will influence the future protein language model community and contribute positively to human health, environmental development, and biomedicine.

\section*{CRediT authorship contribution statement}
Yang Tan: Conceptualization of this study, Methodology, Data curation, Implementation, writing \& editing. Mingchen Li: Conceptualization of this study, Methodology, Data curation, Implementation, writing \& editing. Pan Tan: Review \& editing. Ziyi Zhou: Review \& editing. Huiqun Yu: Supervision, review \& editing. Guisheng Fan: Supervision, review \& editing. Liang Hong: Supervision, review \& editing.

\section*{Declaration of competing interest}
The authors declare that they have no known competing financial interests or personal relationships that could have appeared to influence the work reported in this paper. 

\section*{Data availability}
Data released on GitHub. \url{https://github.com/ginnm/ProteinPretraining}

% \section*{Acknowledgments}
% Supported by Research Programme of National Engineering Laboratory for Big Data Distribution and Exchange Technologies, Shanghai Municipal Special Fund for Promoting High Quality Development (No. 2021-GYHLW-01007)

\bibliography{sample}

\end{document}